\documentclass{bmvc2k}

\title{VESICLE:  Volumetric Evaluation of Synaptic Interfaces using Computer Vision at Large Scale}

\addauthorTwo{William Gray Roncal}{wgr@jhu.edu}{1}{2}
\addauthor{Michael Pekala}{mike.pekala@jhuapl.edu}{2}
\addauthor{Verena Kaynig-Fittkau}{vkaynig@seas.harvard.edu}{3}
\addauthor{Dean M Kleissas}{dean.kleissas@jhuapl.edu}{2}
\addauthorTwo{Joshua T Vogelstein}{jovo@jhu.edu}{4}{5}
\addauthor{Hanspeter Pfister}{pfister@seas.harvard.edu}{3}
\addauthor{Randal Burns}{randal@cs.jhu.edu}{1}
\addauthor{R Jacob Vogelstein}{jacobv@gmail.com}{2}
\addauthor{Mark A Chevillet}{mark.chevillet@jhuapl.edu}{2}
\addauthor{Gregory D Hager}{hager@cs.jhu.edu}{1}

\addinstitution{
Johns Hopkins University, \\ Department of Computer Science, \\ Baltimore, Maryland
}
\addinstitution{
JHU Applied Physics Laboratory, \\
Research and Exploratory Development, \\ Laurel, Maryland
}
\addinstitution{
Harvard University, \\ School of Engineering and Applied Sciences, \\ Cambridge, Massachusetts
}
\addinstitution{
Johns Hopkins University, \\ Department of Biomedical Engineering, \\ Baltimore, Maryland
}
\addinstitution{
Johns Hopkins University, \\ Institute for Computational Medicine, \\ Baltimore, Maryland
}

\runninghead{Gray Roncal et al.}{VESICLE - Scalable Synapse Detection}

\usepackage{stfloats}
\usepackage{placeins}

\newcommand{\website}{\url{openconnecto.me/vesicle}}

\begin{document}

\maketitle

\begin{abstract} 

An open challenge at the forefront of modern neuroscience is to obtain a comprehensive mapping of the neural pathways that underlie human brain function; an enhanced understanding of the wiring diagram of the brain promises to lead to new breakthroughs in diagnosing and treating neurological disorders. Inferring brain structure from image data, such as that obtained via electron microscopy (EM), entails solving the problem of identifying biological structures in large data volumes.  Synapses, which are a key communication structure in the brain, are particularly difficult to detect due to their small size and limited contrast.  Prior work in automated synapse detection has relied upon time-intensive, error-prone biological preparations (isotropic slicing, post-staining) in order to simplify the problem.

This paper presents VESICLE, the first known  approach designed for mammalian synapse detection in anisotropic, non-poststained data.  Our methods explicitly leverage biological context, and the results exceed existing synapse detection methods in terms of accuracy and scalability.  We provide two different approaches - a deep learning classifier (VESICLE-CNN) and a lightweight Random Forest approach (VESICLE-RF), to offer alternatives in the performance-scalability space.  Addressing this synapse detection challenge enables the analysis of high-throughput imaging that is soon expected to produce petabytes of data, and provides tools for more rapid estimation of brain-graphs. Finally, to facilitate community efforts, we developed tools for large-scale object detection, and demonstrated this framework to find $\approx$ 50,000 synapses in 60,000 $\mu m ^3$ (220 GB on disk) of electron microscopy data. \end{abstract}

\section{Introduction}
\label{sec:intro}

Mammalian brains contain billions to trillions of interconnections (i.e., synapses).  To date, the full reconstruction of the neuronal connections of an organism, a ``connectome," has only been completed for nematodes with hundreds of neurons and thousands of synapses  \cite{White1986,Bumbarger2013}.  It is generally accepted \cite{Lichtman2011} that such wiring diagrams are useful for understanding brain function and contributing to medical advances.  For example, many psychiatric illnesses, including autism and schizophrenia, are thought to be ``connectopathies," where inappropriate wiring mediates pathological behavior \cite{Fitzsimmons2013}.  Reliably and automatically identifying synaptic connections (i.e., brain graph edges) is an essential component in understanding brain networks.

Although the community has made great progress towards automatically and comprehensively tracking all neuron fragments through dense electron microscopy data \cite{Gelbart2011,Funke2012}, current state-of-the-art methods for finding synaptic contacts are still insufficient, especially for large-scale automated circuit reconstruction.

In order to detect synapses in electron microscopy data, neuroscientists typically choose to image at $\approx$ 5 nm per voxel in plane, with a slice thickness of $\approx$ 5-70 nm. Capturing complete neurons therefore requires processing terabytes to petabytes of imaged tissue. The largest datasets currently available (and of sufficient size to begin estimating graphs) are acquired using scanning electron microscopy (SEM) or transmission electron microscopy (TEM) due to their high throughput capability \cite{Bock2011,Kasthuri2009}.  These methodologies scale well, but provide a challenging environment for object detection.  The slices are thick relative to in-plane resolution (i.e., anisotropic), due to methodological limitations, and often do not have optimal staining to visually enhance synaptic contacts.  The detection algorithms proposed in this paper are specifically implemented to address these challenges. 
We train a random forest classifier  (VESICLE-RF), which leverages biological context by restricting detections to voxels that have a high probability of being membrane.  This classifier also relies on the identification of neurotransmitter-containing vesicles, which are present near chemical synapses in mammalian brains.  We also present a deep learning classifier (VESICLE-CNN) to find synapses, which has improved performance at the expense of additional computational complexity.  

Both of the VESICLE classifiers provide state-of-the-art performance, and users may choose either method, depending on their environment (e.g., the importance of performance vs. run time, computational resources, availability of human proofreaders).  Our classifiers provide new opportunities to assess neuronal connectivity and can be extended to other datasets and environments.  
%
%

\subsection{Previous Work}
\begin{figure}[ht]
\centering
\includegraphics[width=\textwidth]{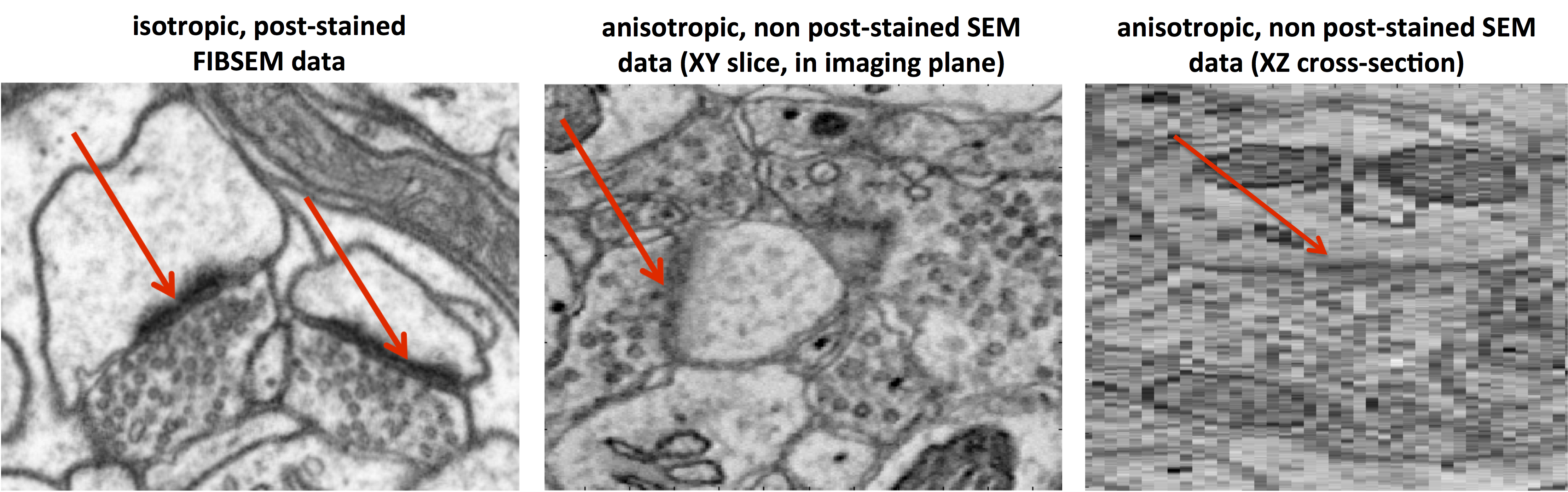}
\caption{Previous work on synapse detection has focused on isotropic post-stained data (left), which shows crisp membranes and dark fuzzy post synaptic densities (arrows) from all orientations. The alternative imaging technique of non post-stained, anisotropic data (middle, right) promises higher throughput, lack of staining artifacts, reduction in lost slices, and less demanding data storage requirements - all critically important for high-throughput connectomics.  The XZ plane of a synapse in anisotropic data is shown (right), illustrating the effect of lower resolution.  We address this more challenging environment, in which membranes appear fuzzier and are harder to distinguish from synaptic contacts. Data courtesy of Graham Knott (left) and Jeff Lichtman (middle, right).} 
%
%
\label{fig:canonical}
\end{figure}

Previous methods for synapse detection have taken several approaches, including both manual and automated methods. Two recent approaches, Kreshuk2011 \cite{Kreshuk2011} and Becker2013 \cite{Becker2013} address the synapse detection problem in post-stained, isotropic, focused ion beam scanning electron microscopy (FIBSEM) data.  \\Kreshuk2011 uses a Random Forest voxel-based classifier and texture-based features to identify pronounced post-stained post-synaptic densities.  This approach is insufficient for our application because of the anisotropy and much lower contrast of our synaptic regions (see Figure~\ref{fig:canonical}), as well as the computational expense.  Becker2013 also uses a voxel-based classification approach and features similar to Kreshuk2011.  However, Becker2013 extends the approach by considering biological context from surrounding pre- and post-synaptic regions at various sizes and locations, based on the synapse pose.  This technique relies on full 3D contextual information and greatly reduces false positives compared to  Kreshuk2011.  

Our result was directly compared to the Becker2013 method \cite{Becker2013} (which was found to be superior to Kreshuk2011 \cite{Kreshuk2011}).  Other work on synapse detection exists but was not used as a comparison method in this manuscript; some methods rely on post-stained data \cite{Kreshuk2014}, post-stained data and accurate cell segmentation \cite{Mishchenko2010}, or post-staining and tailoring for Drosophila (fly) synapses, which have a very different appearance \cite{Plaza2014}.



\section{Methods}

\subsection{Biological Context}
Synapses occur along cell membrane boundaries, between (at least) two neuronal processes.  Although synapses occur in many different configurations, the majority of connections annotated in this dataset are axo-dendritic connections.  The pre-synaptic axonal side is known as a bouton, characterized by a bulbous end filled with small, spherical vesicles.  The synaptic interface is often characterized by a roughly ellipsoidal collection of dark, fuzzy voxels.  In VESICLE-RF, we attempt to directly capture these features. 

\begin{figure}[h!]
\centering
\includegraphics[width=\textwidth]{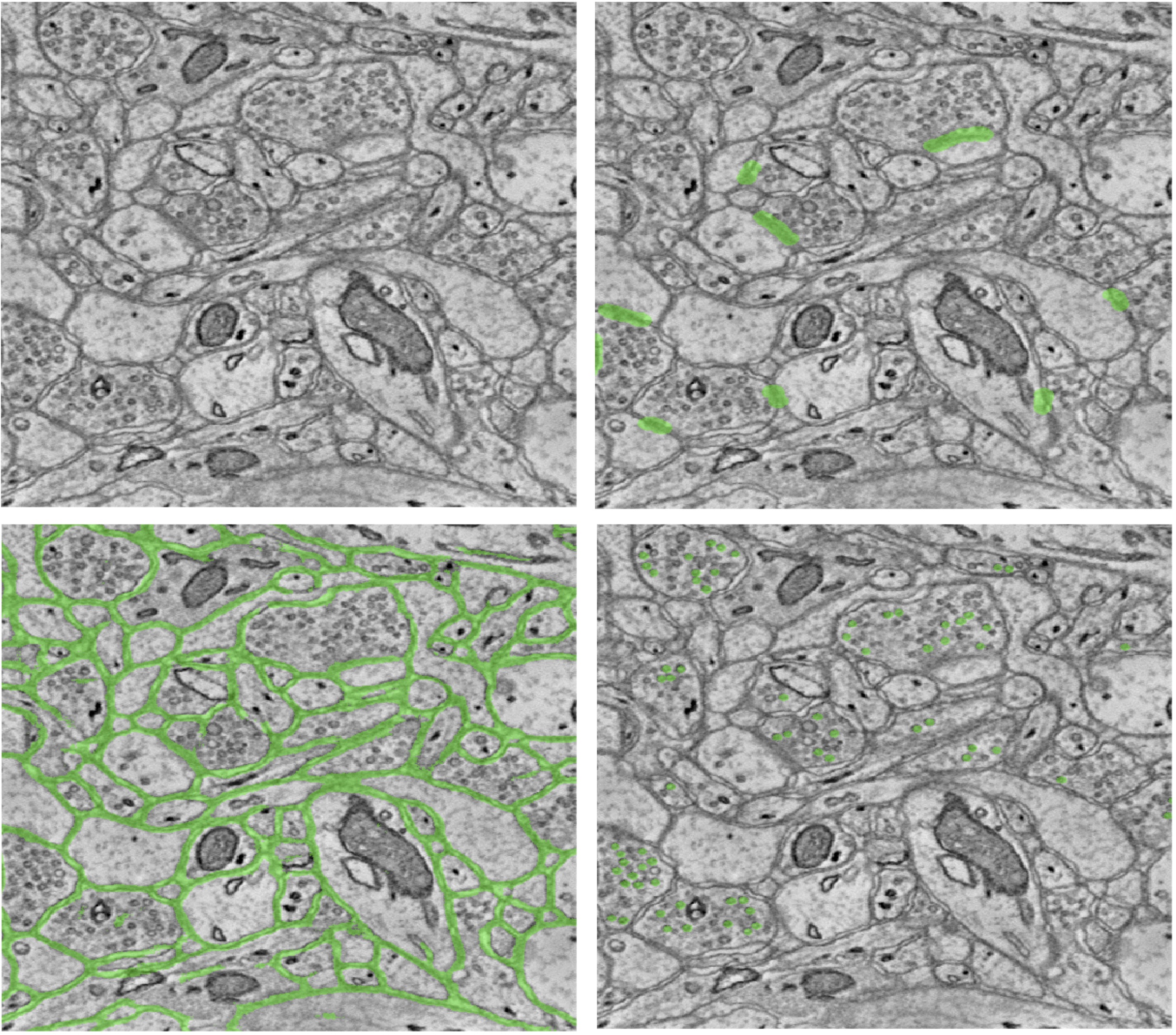}
\caption{A single cross-section of EM data is shown (upper left). The detection task is to identify synapses shown in green (upper right).  These synapses are known to exist at the interface of two neurons; these boundaries can be approximated by previously computed membranes, allowing us to restrict the evaluation regions to the green pixels (lower left).  Clusters of vesicles are a good indicator of an axonal bouton, suggesting that one or more synaptic sites is likely nearby. Vesicles found by our automated detection step are highlighted in green (lower right).}
%
%
\label{fig:bioprior}
\end{figure}

Prior to feature extraction, we leverage membranes (found using the deep learning approach \cite{ciresan2012deep}) which greatly reduces the computational burden and provides a more targeted learning environment for the classifier (Figure~\ref{fig:bioprior}).  The membrane-finding step is computationally intensive (requiring about 3 weeks on 27 Titan GPU cards); however current approaches to neuron detection require membrane probabilities (e.g., \cite{Nunez-Iglesias2013,Kaynig2013d}) and so this is a sunk cost that leverages previously computed information. 

We also identify clusters of vesicles by finding maximal responses to a matched filter extracted from real data followed by clustering to suppress false positives.  This  detector acts as a putative bouton feature to localize regions containing \\synapses.  The vesicles are also of biological interest (e.g., for synapse strength estimation).   Vesicle detection is very lightweight and contributes negligibly to total run time (requiring only 3 hours for the entire 60,000 $\mu m^3$ evaluation volume).

%
%

\subsection{Random Forest Context Aware Classifier (VESICLE-RF)} \label{sec:RF}
We opted for a random forest classifier \cite{Breiman2001} due to its robust finite sample performance in relatively high-dimensional and nonlinear settings.  Furthermore, recent research suggests that this approach significantly outperforms other methods on a variety of tasks \cite{Caruana2008,Fernandez-Delgado2014}.  The output of the random forest is a scalar probability for each pixel, which we threshold and post-process to obtain a class label, as described below.  


We began with a large set of potential feature descriptors (e.g., Haralick features, Gabor wavelets, structure tensors), and evaluated their performance based on a combination of Random Forest importance on training and validation data, computational efficiency, and ability to capture biologically significant characteristics.  We pruned this feature set down to ten features, retaining state-of-the-art performance in a computationally lightweight package.  For efficiency reasons, we computed features in a two-pass approach.  We first computed several data transforms on the two-dimensional EM slice data (to better account for the image anisotropy).  We then created our features by convolving box kernels of different bandwidths with the results from the previous step.  This allows information to be summarized at different scales, as explained further in Table~\ref{features}.  Finally, we computed a feature capturing the minimum distance to a neurotransmitter-containing vesicle. 

\begin{table}[ht]
\caption{Description of features used in VESICLE-RF; data transforms are summarized using different kernel bandwidths:  $\theta_0:  [5,5,1]$, $\theta_1: [15,15,3]$, $\theta_2 = [25,25,5]$, $\theta_3 = [101,101,5]$, $\theta_4 =$  minimum vesicle distance.} \label{features}
\centering
\begin{tabular}{| p{4cm} | p{6.5cm} |}
\hline
\textbf{Data Transform}& \textbf{Box Kernel}\\
\hline\hline
Intensity & $\theta_0, \theta_1$\\\hline
Local Binary Pattern & $\theta_0$\\\hline
Image Gradient Magnitude& $\theta_1, \theta_2$\\\hline
Vesicles & $\theta_2, \theta_3, \theta_4$\\\hline
Structure Tensor & $\theta_1, \theta_2$\\\hline
\end{tabular}

\end{table}

We train our classifier using 200,000 samples (balanced synapse, non-synapse classes).  Putative synapse candidates are fused into 3D objects by thresholding and size filtering as described in section~\ref{sec:perfeval}. This method is scalable, requiring only a small amount of computational time and resources to train and test.

\subsection{Deep Learning Classifier - VESICLE-CNN}

Deep convolutional neural networks (CNNs) have recently provided state-of-the-art performance across a wide range of image and video recognition problems.  These successes include a number of medical imaging applications; a small sample includes mitosis detection \cite{cirecsan2013mitosis}, organ segmentation \cite{roth2015deep} and membrane detection in EM data \cite{ciresan2012deep}.  
The recent success in membrane detection is particularly compelling given the common imaging modality and visual similarity between membranes and synapses.
To test the hypotheses that CNNs may also provide an effective means of classifying synapses we adopt and re-implement the pixel-level classification approach of \cite{ciresan2012deep} suitably adapted for our application.

As in section~\ref{sec:RF}, each pixel in the EM cube is presumed to be either a synapse pixel or a non-synapse pixel.  The features used for classification consist of a $65 \times 65$ tile centered on the pixel location of interest.  
For classification we use a CNN with three convolutional layers and two fully connected layers, roughly corresponding to the CNN designated ``N3" in \cite{ciresan2012deep}. This CNN is implemented using the Caffe deep learning framework \cite{Jia2014}; the full architecture specification (e.g. types of nonlinearites and specific layer parameters) is encoded in the Caffe configuration files which are provided as part of our open source code.

During training we balance the synapse (target) and non-synapse (clutter) examples evenly; since synapse pixels are relatively sparse this involves substantially subsampling the majority class.  To focus the training on examples that are presumably the most challenging, we threshold the membrane probabilities described above, and use the result as a bandpass filter.  Negative examples are drawn randomly from the set of non-synapse membrane pixels.  We also add synthetic data augmentation by rotating the tiles in each mini-batch by a random angle (the insight behind this step is that synapses may be oriented in any direction).  Our test paradigm does not rely on membrane probabilities; once trained, the deep learning classifier requires only EM data as input.  The neurotransmitter-containing vesicles used in VESICLE-RF are not used in VESICLE-CNN for training or test.

\subsection{Scalable Processing}
Vision processing of large neuroscience data requires unique and well-designed infrastructure; our approach is designed to eventually process petabytes of data. Specifically, scalable computer vision requires storage and retrieval of images, annotation standards for semantic labels, and a distributed processing framework to process data and perform inference across blocks that are too large to fit in RAM on a commodity workstation.\\
We leverage the Open Connectome Project infrastructure \cite{Burns2013} to store and retrieve data and the Reusable Annotation Markup for Open coNnectomes (RAMON) data model \cite{GrayRoncal2015}.  We also leverage the Laboratory of Neuroimaging (LONI) Pipeline \cite{Rex2003a} workflow manager to develop solutions to object detection challenges inherent in big neuroscience. 

As part of this work, we built a generic object detection pipeline leveraging the Open Connectome project that can be used to train new synapse detection algorithms or detect other targets of interest.  

\section{Results}

Our VESICLE classifiers were trained and evaluated for both performance and scalability, exceeding existing state-of-the-art performance. 

\subsection{Data}

VESICLE was evaluated on an anisotropic ($3\times 3\times 30$ $nm$ resolution) color- corrected \cite{Kazhdan} dataset of non-poststained mouse somatosensory cortex \cite{Kasthuri2009}.  This is the largest known dataset of its kind.  Prior to all processing, we downsampled the data to $6\times 6\times 30$ $nm$ resolution.  The training and test volumes were extracted from this larger EM volume.  For training, each method used a $1024 \times 1024 \times 100$ $\mu m^3$ region of data (denoted AC4).  For testing, a non-contiguous, equally sized cube from the same dataset was evaluated (denoted AC3).  For the deep learning algorithm a different size pad region was used due to training methodologies.  A padded border was used on the test region for all algorithms to ensure that all labeled synapses in the volume were available for evaluation.

Gold standard labels for synapses were provided by expert neurobiologist annotators.  The training labels were assumed to be correct (our classification result was evaluated in an open-loop process). 


\subsection{Performance Evaluation}
\label{sec:perfeval}
We assess our performance by evaluating the precision-recall of synaptic objects.  Pixel error, while potentially useful for characterizing synaptic weight and morphology, is a less urgent goal for connectomics, which must first identify the connections between neurons before ascribing attributes.  A focus on pixel accuracy also can obscure the actual task of connection detection.

\begin{figure}[ht]
    \centering
    \includegraphics[width=\linewidth]{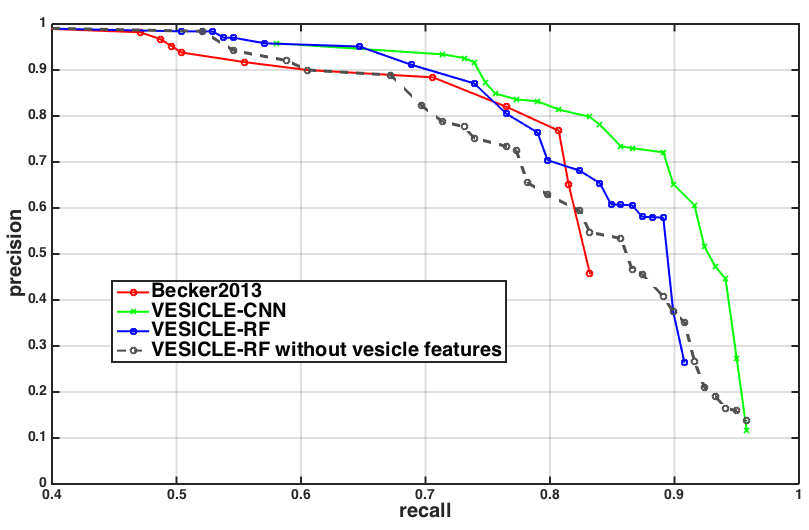}
    
    \caption{VESICLE-RF and VESICLE-CNN significantly outperform prior state-of-the art, particularly at high recall rates.  The relatively abrupt endpoint of the Becker2013 method occurs because beyond this point, thresholded probabilities group into large detected regions rather than individual synapses, which are disallowed (as described above).
} 
\label{fig:resultspr}
\end{figure}

A quantitative comparision of our performance relative to existing work is presented in Figure~\ref{fig:resultspr}.  Of particular significance is our performance at high recall operating points.  For many connectomics applications, it is essential to ensure that the majority of connections are captured (i.e., low false negative rate); false positives can be remediated through a variety of approaches (e.g., biological plausiblity based on incident neurons, manual proofreading).  

To construct precision-recall curves from VESICLE-RF and VESICLE-CNN pixel-level classification results, we developed a procedure to sweep over probability score thresholds (0.5-1.0) and create initial objects through a connected component analysis.  We generated additional operating points by varying biologically motivated size (2D:  0-200 minimum, 2500-10000 maximum; 3D:  100-2000 minimum) and slice persistence (1-5 slices) requirements.    For Becker2013, we ran the statically linked package provided by the author on our data volumes.  We then followed their suggested method (similar to the VESICLE approach) to create synaptic objects from raw pixel probabilities by thresholding probabilities (0.0-1.0), running a connected component algorithm and rejecting all objects comprised of fewer than 1000 voxels \cite{Becker2013}.

When computing object metrics, we computed true positives, false positives, and false negatives by examining overlapping areas between truth labels and detected objects.  We added the additional constraint of allowing each detection to count for only one truth detection to disallow large synapse detections that cover many true synapses and provide little intuition into connectomics questions. A version of the classifier was trained without the vesicle features to provide insight into the importance of biological context in this problem domain.

\begin{figure*}[ht]
\centering
\includegraphics[width=\textwidth]{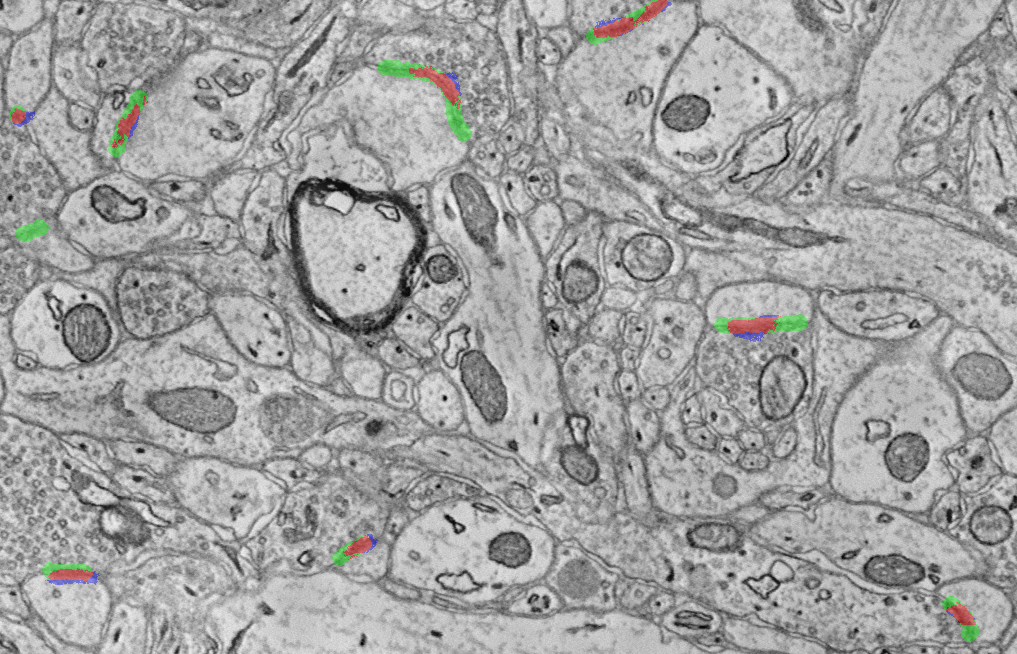}
\caption{Example VESICLE result.  Gold standard labels are shown in green, and VESICLE-RF detections are shown in blue.  Red pixels represent True Positives (TP).  Objects that are only green are False Negatives (FN) and objects that are only blue are False Positives (FP).  Object detection results are analyzed in 3D and so single slices may be misleading.}
\label{fig:detection}
\end{figure*}

A qualitative visualization of our VESICLE-RF performance is shown in Figure~\ref{fig:detection}.   

%
%
%
%
%

\subsection{Scalability Analysis}
VESICLE-RF enables large-scale processing because of its light computational footprint and ability to be easily parallelized in a High Performance Computing (HPC) CPU environment.  Relative to Becker2013 \cite{Becker2013}, this approach is dramatically less computationally intensive for training (8 GB RAM, 10 minutes v. 20 GB RAM, 11 hours). During evaluation our approach is approximately twice as fast (10 minutes versus 20,  using unoptimized MATLAB code), and has one-eighth the maximum computational load and half the maximum RAM requirement.  VESICLE-CNN required  56 hours to train and 39 hours to evaluate the test cuboid on a single GPU.

\begin{figure*}[h]
\centering
\includegraphics[width=\textwidth]{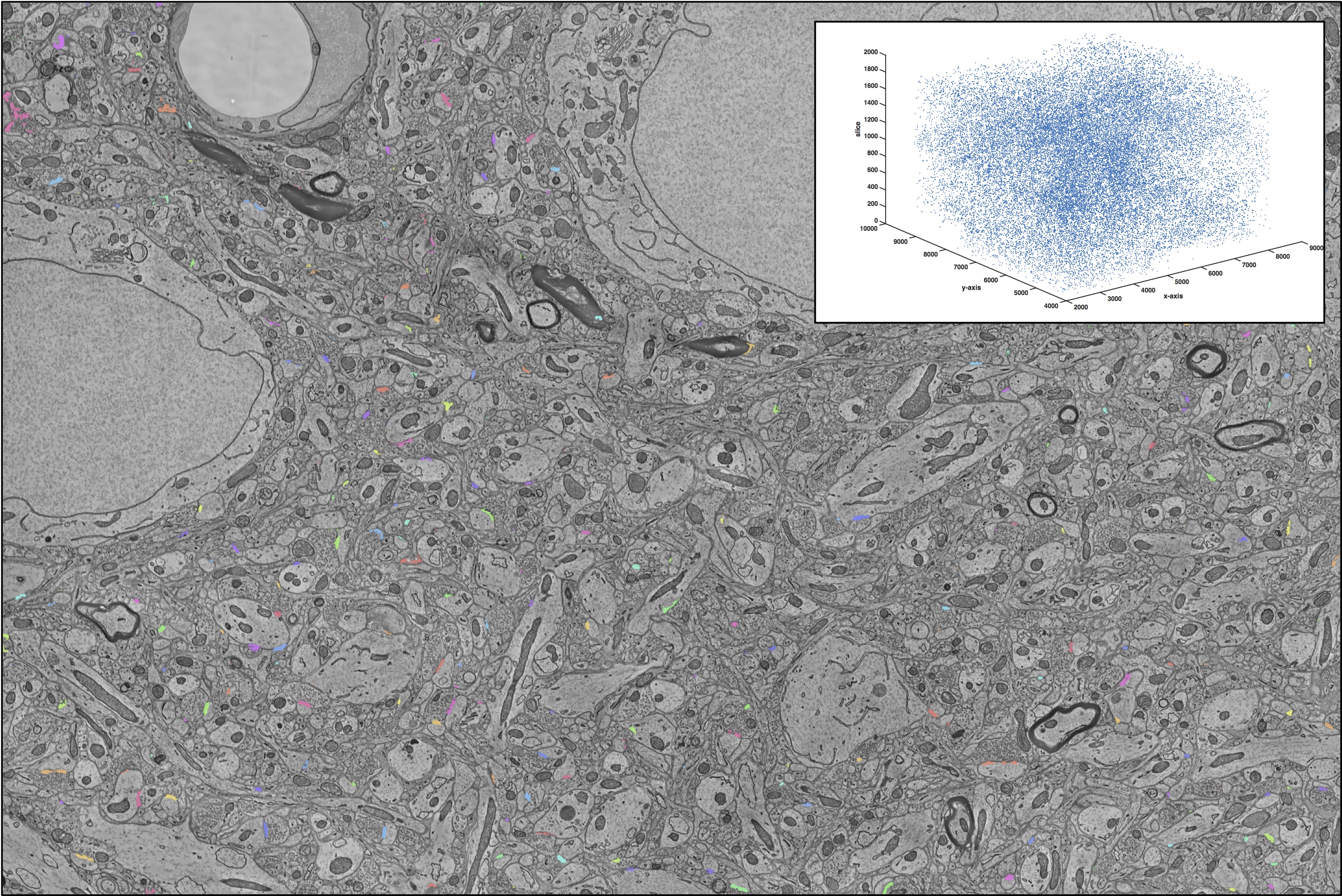}
\caption{Visualization of large scale synapse detection results; we found a total of 50,000 putative synapses in our volume.  An XY slice showing detected synapses is shown, and a point cloud of the synapse centroids are also visualized (inset). A full resolution version of this image is available via RESTful query as explained here: \url{http://openconnecto.me/vesicle}.  Each synapse is represented by a different color label.}

\label{fig:large_scale_results}
\end{figure*}

To demonstrate the scalability of our approach and our distributed processing framework, we applied our VESICLE-RF classifier to the largest available non-poststained, anisotropic dataset \cite{Kasthuri2009}.  The inscribed cuboid is $\approx 220$ GB on disk; we downsample by a factor of two in the X and Y dimensions prior to processing (60,000 $\mu m^3$, 56GB on disk after downsampling).   In Figure~\ref{fig:large_scale_results}, we show a visualization of the 50,335 synapses found in this analysis.  We chose the VESICLE-RF method here to emphasize the advantages of scalable classifiers; this method is  $\approx 200$ times faster than VESICLE-CNN (evaluated as a single job on the same data cube).  When deploying our classifier at scale, we increase our pad size to be larger than a synaptic cleft, and discard border detections; this allows us to avoid boundary merge issues.  We also optionally allow the detection threshold to vary based on pixel probabilities to improve robustness.


Although we focus on the problem of non-poststained data, we provide additional evidence of scalability by running on an anisotropic post-stained dataset \cite{Bock2011}.  This dataset is 20 teravoxels on disk (14.5 million $\mu m^3$). We downsampled by a factor of two in the X and Y dimensions, and ran a variant of our detector on 5 TB of data. We found approximately 11.6 million synapses, consistent with published estimates of synapse density  (0.5-1 synapse/$\mu m^3$)\cite{Busse2013}. Although our assessment is ongoing due to limited available truth, we estimate an operating point at a precision of 0.69 and a recall of 0.6, using the available labels.  Because this data was poststained, we estimated membrane detections by applying an intensity-based bandpass filter (with cutoffs derived as the [0.02-0.80] values of the synapse training data labels).  We expect that performance will improve with additional development.  

\section{Discussion}
In this paper we have presented two algorithms for synapse detection in non- poststained, anisotropic EM data, and have shown that both perform better than state-of-the-art methods.  The Random Forest approach offers a scalable solution with an approach inspired by expert human annotators, while the deep learning result achieves the best overall performance. 

We built and demonstrated a reusable, scalable pipeline using the Open Connectome Project services and used it to find putative synapses on large cubes of mammalian EM data.  We presented the largest result known by orders of magnitude (in both volume processed and synaptic detections).  As future work, we plan to refine our estimates of synapse morphology and position by implementing a region-growing algorithm and incorporating additional contextual information. We also plan to utilize supervoxel methods in both of our approaches, and consider VESICLE-RF texture features inspired by CNN results.

For the VESICLE-CNN approach, future work includes exploring alternative CNN architectures (such as the very deep networks considered in \cite{Simonyan2015}), enhancing the tile-based input features (e.g. to include three dimensional context) and improving the computational complexity (through the use of sampling techniques guided by our biological priors and/or computational techniques such as those described in \cite{Giusti2013}).

Our code and data are open source and available at: \website.

%

\section{Acknowledgements}

The authors thank Bobby Kasthuri, Daniel Berger, and Jeff Lichtman for providing electron microscopy data and truth labels, and Lindsey Fernandez for helpful discussions on algorithm development and comparison metrics. This work is partially supported by JHU Applied Physics Laboratory Internal Research and Development funds, by NIH NIBIB 1RO1EB016411-01, and by NSF grants OIA-1125087 and IIS 1447344. 

\FloatBarrier

\bibliographystyle{bmvc2k}
\bibliography{vesicle.bib}	

\end{document}